\documentclass[a4paper]{article}

\usepackage{acl2010}
\usepackage{latexsym}
\usepackage{url}
\usepackage{graphicx}
\usepackage{picture}
\usepackage{color}
\usepackage{amsmath}
\usepackage{amssymb}
\usepackage{MnSymbol}
\usepackage{ifsym}

\title{KSU KDD: Word Sense Induction by Clustering in Topic Space}
\author{Wesam Elshamy, Doina Caragea, William H. Hsu\\ Kansas State University\\ \texttt{\{welshamy, dcaragea, bhsu\}@ksu.edu}}

\begin{document}
\maketitle

\begin{abstract}
We describe our language-independent unsupervised word sense induction system.  This system only uses topic features to cluster different word senses in their global context topic space.  Using unlabeled data, this system trains a latent Dirichlet allocation (LDA) topic model then uses it to infer the topics distribution of the test instances.  By clustering these topics distributions in their topic space we cluster them into different senses.  Our hypothesis is that closeness in topic space reflects similarity between different word senses.  This system participated in SemEval-2 word sense induction and disambiguation task and achieved the second highest V-measure score among all other systems.
\end{abstract}

\section{Introduction}
Ambiguity of meaning is inherent in natural language because the deliverer of words tries to minimize the size of the vocabulary set he uses.  Therefore, a sizable portion of this vocabulary is polysemous and the intended meaning of such words can be encoded in their context.

Due to the knowledge acquisition bottleneck problem and scarcity in training data \cite{cai07}, unsupervised corpus based approaches could be favored over supervised ones in word sense disambiguation (WSD) tasks.

Similar efforts in this area include work by Cai et al. \cite{cai07} in which they use latent Dirichlet allocation (LDA) topic models to extract the global context topic and use it as a feature along other baseline features.  Another technique uses clustering based approach with WordNet as an external resource for disambiguation without relying on training data \cite{henry07}.

To disambiguate a polysemous word in a text document, we use the document topic distribution to represent its context.  A document topic distribution is the probabilistic distribution of a document over a set of topics.  The assumption is that: given two word senses and the topic distribution of their context, the closeness between these two topic distributions in their topic space is an indication of the similarity between those two senses.

Our motivation behind building this system was the observation that the context of a polysemous word helps determining its sense to some degree.  In our word sense induction (WSI) system, we use LDA to create a topic model for the given corpus and use it to infer the topic distribution of the documents containing the ambiguous words.

This paper describes our WSI system which participated in SemEval-2 word sense induction and disambiguation task \cite{manandhar10}.

\section{Latent Dirichlet allocation}
\begin{figure}
  \centering
  \resizebox{0.8\linewidth}{!}{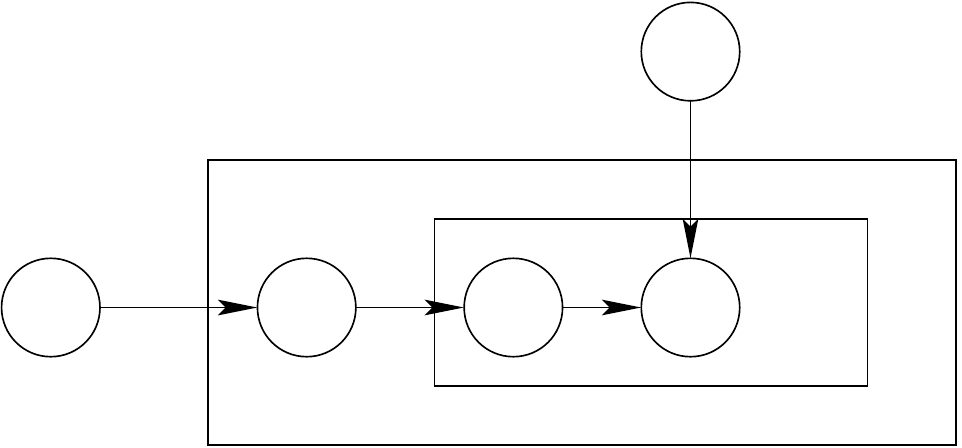}
  \caption{A graphical model for LDA\label{lda}}  
\end{figure}

LDA is a probabilistic model for a collection of discrete data \cite{blei03}.  It can be graphically represented as shown in Figure \ref{lda} as a three level hierarchical Bayesian model.  In this model, the corpus consists of $M$ documents, each is a multinomial distribution over $K$ topics, which are in turn multinomial distributions over words.

To generate a document $d$ using this probabilistic model, a distribution over topics $\theta_d$ is generated using a Dirichlet prior with parameter $\alpha$.  Then, for each of the $N_d$ words $w_{dn}$ in the document, a topic $z_{dn}$ is drawn from a multinomial distribution with the parameter $\theta_d$.  Then, a word $w_{dn}$ is drawn from that topic's distribution over words, given  $\beta_{ij} = p(w=i|z=j)$.  Where $\beta_{ij}$ is the probability of choosing word $i$ given topic $j$.

\section{System description}
We wanted to examine the trade-off between simplicity, cost and performance by building a simple language-independent, totally unsupervised, computationally cheap system and compare its performance to other WSI systems participating in the SemEval-2 WSI task \cite{manandhar10}.  We expect a degradation in precision of our simple approach as the granularity of senses becomes finer;  This is due to the degrading sensitivity in mapping between the topics space and the senses space.  We note that our simple approach will fail if multiple senses of the same word appear in the same document;  Since these senses will be represented by the same topic distribution of the document, they will be clustered in the same cluster.

Our system is a language-independent system.  The used LDA topic model has no knowledge of the training or testing corpus language.  Unlike most other WSI and WSD systems, it doesn't make use of part of speech (POS) features which are language dependent and require POS annotated training data.  The only features used are the topics distribution of bag-of-words containing the ambiguous word.

First, for each target polysemous word $wp$ (noun or verb), we train a MALLET\footnote{http://mallet.cs.umass.edu} parallel topic model implementation of LDA on all the training instances of that word.  Then we use the trained topic model to infer the topics distribution $\theta_l$ for each of the test instances of that word.  For a $K$-topics topic model, each topics distribution can be represented as a point in a $K$-dimensional topic space.  These points can be clustered into $C$ different clusters, each representing a word sense.  We used MALLET's $K$-means clustering algorithm with cosine similarity to measure the distance between different topic distributions in the topic space.

\section{Evaluation measures}
We use the same unsupervised evaluation measures used in SemEval-2 \cite{manandhar09}.  These measures do not require descriptive 

The V-measure is used for unsupervised evaluation.  It is the harmonic mean of the \emph{homogeneity} and \emph{completeness}.  Homogeneity is a measure of the degree that each formed cluster consists of data points that belong to a single gold standard (GS) class as defined below.

\begin{gather}
  homogeneity = 1 - \frac{H(GS|C)}{H(GS)}\\
  H(GS) = -\sum_{i=1}^{|GS|}\frac{\sum_{j=1}^{|C|}a_{ij}}{N}\log\frac{\sum_{j=1}^{|C|}a_{ij}}{N} \label{hgs}\\
  H(GS|C) = -\sum_{j=1}^{|C|}\sum_{i=1}^{|GS|}\frac{a_{ij}}{N}\log\frac{a_{ij}}{\sum_{k=1}^{|GS|}a_{kj}}\label{hgsc}
\end{gather}

Where $H()$ is an entropy function, $|C|$ and $|GS|$ refer to cluster and class sizes, respectively.  $N$ is the number of data points, $a_{ij}$ are data points of class $GS_i$ that belong to cluster $C_j$.

On the other hand, completeness measures the degree that each class consists of data points that belong to a single cluster.  It is defined as follows.

\begin{gather}
  completeness = 1 - \frac{H(C|GS)}{H(C)}\\
  H(C) = -\sum_{j=1}^{|C|}\frac{\sum_{i=1}^{|GS|}a_{ij}}{N}\log\frac{\sum_{i=1}^{|GS|}a_{ij}}{N} \label{hc}\\
  H(C|GS) = -\sum_{i=1}^{|GS|}\sum_{j=1}^{|C|}\frac{a_{ij}}{N}\log\frac{a_{ij}}{\sum_{k=1}^{|C|}a_{ik}}\label{hcgs}
\end{gather}

Homogeneity and completeness can be seen as entropy based measures of precision and recall, respectively.  The V-measure has a range of 0 (worst performance) to 1, inclusive.

The other evaluation measure is the F-score, which is the harmonic mean of precision and recall.  It has a range of 0 to 1 (best performance), inclusive.

\section{Experiments and results}
The WSI system described earlier was tested on SemEval-1 WSI task (task 2) data (65 verbs, 35 nouns), and participated in the same task in SemEval-2 (task 14) (50 verbs, 50 nouns).  The sense induction process was the same in both cases.

Before running our main experiments, we wanted to see how the number of topics $K$ used in the topic model could affect the performance of our system.  We tested our WSI system on SemEval-1 data using different $K$ values as shown in Table \ref{keffect}.  We found that the V-measure and F-score values increase with increasing $K$, as more dimensions are added to the topic space, the different senses in this $K$-dimensional space unfold.  This trend stops at a value of $K=400$ in a sign to the limited vocabulary of the training data.  This $K$ value is used in all other experiments.

\begin{table}
  \centering
  \caption{Effect of varying the number of topics $K$ on performance\label{keffect}}
  \vspace{0.5em}
  \begin{tabular}{lccccc}
    K         &10   &50  &200  &400  &500\\
    \hline
    V-measure &5.1  &5.8  &7.2  &8.4  &8.1\\
    F-score   &8.6  &32.0 &53.9 &63.9 &64.2\\
    \hline 
  \end{tabular}
\end{table}

Next, we evaluated the performance of our system on SemEval-1 WSI task data.  Since no training data was provided for this task, we used an un-annotated version of the test instances to create the LDA topic model.  For each target word (verb or noun), we trained the topic model on its given test instances.  Then we used the generated model's inferencer to find the topics distribution of each one of them.  These distributions are then clustered in the topic space using the $K$-means algorithm and the cosine similarity measure was used to evaluate the distances between these distributions.  The results of this experiment are shown in Table \ref{vfsemeval1}.

\begin{table}
  \centering
  \caption{V-measure and F-score on SemEval-1\label{vfsemeval1}}
  \vspace{0.5em}
  \begin{tabular}{lccc}
              &All  &Verbs &Nouns\\
    \hline
    V-measure &8.4  &8.0   &8.7\\
    F-score   &63.9 &56.8  &69.0\\
    \hline
  \end{tabular}
\end{table}

Our WSI system took part in the main SemEval-2 WSI task (task 14).  In the unsupervised evaluation, our system had the second highest V-measure value of 15.7 for all words\footnote{A complete evaluation of all participating systems is available online at: \scriptsize{http://www.cs.york.ac.uk/semeval2010\_WSI/task\_14\_ranking.html}}.  A break down of the obtained V-measure and F-scores is shown in Table \ref{vfsemeval2}.  

\begin{table}
  \centering
  \caption{V-measure and F-score on SemEval-2\label{vfsemeval2}}
  \vspace{0.5em}
  \begin{tabular}{lccc}
              &All  &Verbs &Nouns\\
    \hline
    V-measure &15.7 &12.4  &18.0\\
    F-score   &36.9 &54.7  &24.6\\
    \hline
  \end{tabular}
\end{table}

\begin{figure}
  \centering
  \includegraphics[width=\linewidth]{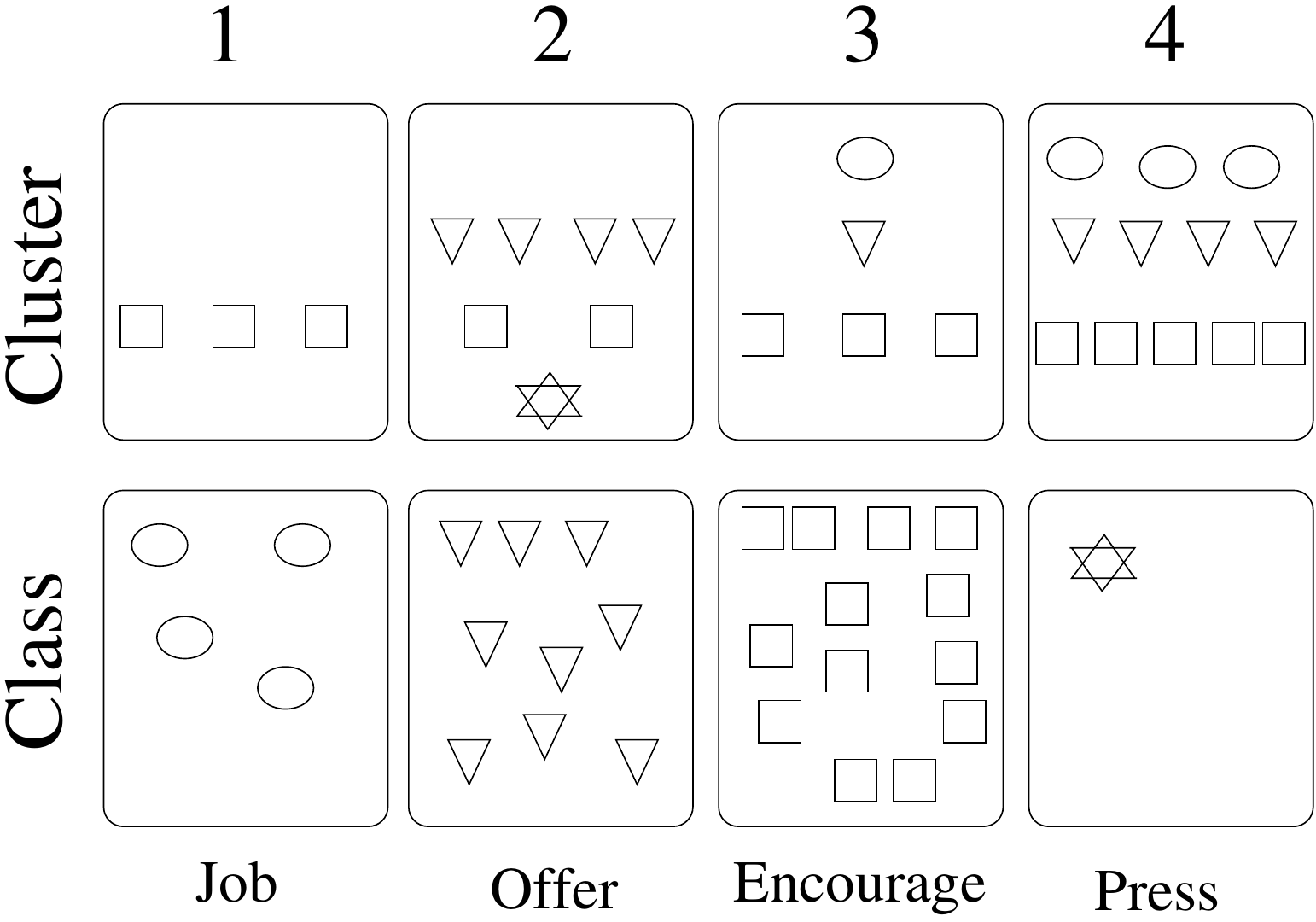}
  \caption{Analysis of sense clustering\label{promo}}
\end{figure}

To analyze the performance of the system, we examined the clustering of the target noun word ``\emph{promotion}'' to different senses by our system.  We compared it to the GS classes of this word in the answer key provided by the task organizers.  For a more objective comparison, we ran the $K$-means clustering algorithm with $K$ equal to the number of GS classes.  Even though the number of formed clusters affects the performance of the system, we assume that the number of senses is known in this analysis.  We focus on the ability of the algorithm to cluster similar senses together.  A graphical comparison is given in Figure \ref{promo}.

The target noun word ``promotion'' has 27 instances and four senses.  The lower four rectangles in Figure \ref{promo} represent the four different GS classes, and the upper four rectangles represent the four clusters created by our system.  Three of the four instances representing a \emph{job} ``promotion'' ($\largecircle$) were clustered together, but the fourth one was clustered in a different class due to terms like ``driving,'' ``troops,'' and ``hostile'' in its context.  The offer sense of ``promotion'' ($\largetriangledown$) was mainly split between two clusters, cluster 2 which most of its instances has mentions of numbers and monetary units, and cluster 4 which describes business and labor from an employee's eye.

The 13 instances of the third class which carry the sense \emph{encourage} of the word promotion ($\largesquare$) are distributed among the four different clusters depending on other topic words that classified them as either belonging to cluster 4 (encouragement in business),  cluster 3 (encouragement in conflict or war context), cluster 2 (numbers and money context), or cluster 1 (otherwise).  We can see that the topic model is unable to detect and extract topic words for the ``encourage'' sense of the word.  Finally, due to the lack of enough training instances of the sense of a promotional issue of a newspaper ($\largestarofdavid$), the topic model inferencer clustered it in the numbers and monetary cluster because it was rich in numbers.

\section{Conclusion}
Clustering the topics distributions of the global context of polysemous words in the topic space to induce their sense is cheap as it does not require any annotated data and is language-independent.

Even though the clustering produced by our system did not fully conform with the set of senses given by the GS classes, it can be seen from the analyzed example given earlier that our clustering carried some different senses.  In one case, a GS sense was not captured by the topic model, and instead, other cues from its instances context were used to cluster them accordingly.  The induced clustering had some noise though.

This simple WSI approach can be used for cheap sense induction or for languages for which no POS tagger has been created yet.  This system which had the second highest V-measure score in SemEval-2 WSI task achieves a good trade-off between performance and cost.

\bibliography{../bibliography}

\begin{thebibliography}{}

\bibitem[\protect\citename{Anaya-S\'{a}nchez \bgroup et al.\egroup
  }2007]{henry07}
Henry Anaya-S\'{a}nchez, Aurora Pons-Porrata, and Rafael Berlanga-Llavori.
\newblock 2007.
\newblock Tkb-uo: Using sense clustering for wsd.
\newblock In {\em Proceedings of the Fourth International Workshop on Semantic
  Evaluations (SemEval-2007)}, pages 322--325, Prague, Czech Republic, June.
  Association for Computational Linguistics.

\bibitem[\protect\citename{Blei \bgroup et al.\egroup }2003]{blei03}
David~M. Blei, Andrew~Y. Ng, and Michael~I. Jordan.
\newblock 2003.
\newblock Latent dirichlet allocation.
\newblock {\em J. Mach. Learn. Res.}, 3:993--1022.

\bibitem[\protect\citename{Cai \bgroup et al.\egroup }2007]{cai07}
Junfu Cai, Wee~Sun Lee, and Yee~Whye Teh.
\newblock 2007.
\newblock Improving word sense disambiguation using topic features.
\newblock In {\em Proceedings of the 2007 Joint Conference on Empirical Methods
  in Natural Language Processing and Computational Natural Language Learning
  (EMNLP-CoNLL)}, pages 1015--1023, Prague, Czech Republic, June. Association
  for Computational Linguistics.

\bibitem[\protect\citename{Manandhar and Klapaftis}2009]{manandhar09}
Suresh Manandhar and Ioannis~P. Klapaftis.
\newblock 2009.
\newblock Semeval-2010 task 14: evaluation setting for word sense induction \&
  disambiguation systems.
\newblock In {\em DEW '09: Proceedings of the Workshop on Semantic Evaluations:
  Recent Achievements and Future Directions}, pages 117--122, Morristown, NJ,
  USA. Association for Computational Linguistics.

\bibitem[\protect\citename{Manandhar \bgroup et al.\egroup }2010]{manandhar10}
Suresh Manandhar, Ioannis~P. Klapaftis, Dmitriy Dligach, and Sameer~S. Pradhan.
\newblock 2010.
\newblock Semeval-2010 task 14: Word sense induction \& disambiguation.
\newblock In {\em Proceedings of SemEval-2}, Uppsala, Sweden. ACL.

\end{thebibliography}
\bibliographystyle{acl}

\end{document}